\documentclass[11pt]{article}

\usepackage{acl2026}

\usepackage{times}
\usepackage{latexsym}

\usepackage[T1]{fontenc}

\usepackage[utf8]{inputenc}
\usepackage{microtype}
\usepackage{inconsolata}

\usepackage{graphicx}
\usepackage{array}
\usepackage{multirow}
\usepackage[most]{tcolorbox}
\usepackage{amsmath} 
\usepackage{booktabs}
\usepackage{mathtools}
\usepackage{tabularx}

\newcommand{\ie}{\emph{i.e.}}
\newcommand{\eg}{\emph{e.g.}}

\def \OURS{\texttt{MTDEval}}
\def \OURDATA{{P$^2$-MTD}}

\title{Learning an Efficient Multi-Turn Dialogue Evaluator from \\Multiple LLM Judges}

\begin{document}



\author{
  Yuqi Tang$^{1,2}$,
  Kehua Feng$^{2,3}$,
  Yunfeng Wang$^{4}$,
  Zhiwen Chen$^{4}$,
  Chengfei Lv$^{4}$,
  Gang Yu$^{4}$, \\
  \textbf{Qiang Zhang}$^{1,2}$,
  \textbf{Keyan Ding}$^{2*}$,
  \textbf{Huajun Chen}$^{2,3}$\thanks{Corresponding author.} \\
  $^1$ZJU--UIUC Institute, Zhejiang University \\
  $^2$ZJU--Hangzhou Global Scientific and Technological Innovation Center, Zhejiang University \\
  $^3$College of Computer Science and Technology, Zhejiang University \\
  $^4$Alibaba Group \\
  \texttt{yuqi.22@intl.zju.edu.cn}, \texttt{\{kehuafeng, dingkeyan\}@zju.edu.cn}
}

\maketitle

\begin{abstract}
Evaluating the conversational abilities of large language models (LLMs) remains a challenging task. Current mainstream approaches primarily rely on the ``LLM-as-a-judge" paradigm, where an LLM is prompted to serve as an evaluator to assess dialogue quality. However, such methods often suffer from various biases, which undermine the reliability and consistency of the evaluation results. To mitigate these biases, recent methods employ multiple LLMs as judges and aggregate their judgments to select the optimal assessment. Although effective, this multi-judge approach incurs significant computational overhead during inference.
In this paper, we propose an efficient dialogue evaluator that captures the collective wisdom of multiple LLM judges by aggregating their preference knowledge into a single model.  Our approach preserves the advantages of diverse multi-judge feedback while drastically reducing the evaluation cost, enabling fast, flexible, and fine-grained dialogue quality assessment. Extensive experiments on seven single rating and pairwise comparison dialogue evaluation benchmarks demonstrate that our method outperforms existing baselines across diverse scenarios, showcasing its efficiency and robustness.

\end{abstract}

\section{Introduction}
Measuring the quality of dialogues generated by large language models (LLMs) presents significant challenges due to the inherent complexity and multi-dimensional nature of dialogue interactions. Recent advancements in LLMs have significantly enhanced their capability to evaluate single-turn dialogues \cite{li2025beyond}, yet assessing LLM performance in multi-turn dialogue settings remains highly challenging, particularly in evaluating critical capabilities such as instruction following, self-coherence, and emotional alignment \cite{sirdeshmukh2025multichallenge}. Traditional automated evaluation metrics (e.g., BLEU \cite{papineni2002bleu}, ROUGE \cite{lin2004rouge}, and BERTScore \cite{zhang2019bertscore}), which rely on fixed lexical or semantic overlaps, often fail to effectively capture the flexibility of natural language as perceived by humans, as well as the rich semantics present in multi-turn dialogues \cite{feng2025samer}. Moreover, these metrics require external references to evaluate knowledge-grounded responses, limiting their applicability in scenarios where such references are unavailable.

\begin{figure}[t]
  \centering
  \includegraphics[width=0.5\textwidth]{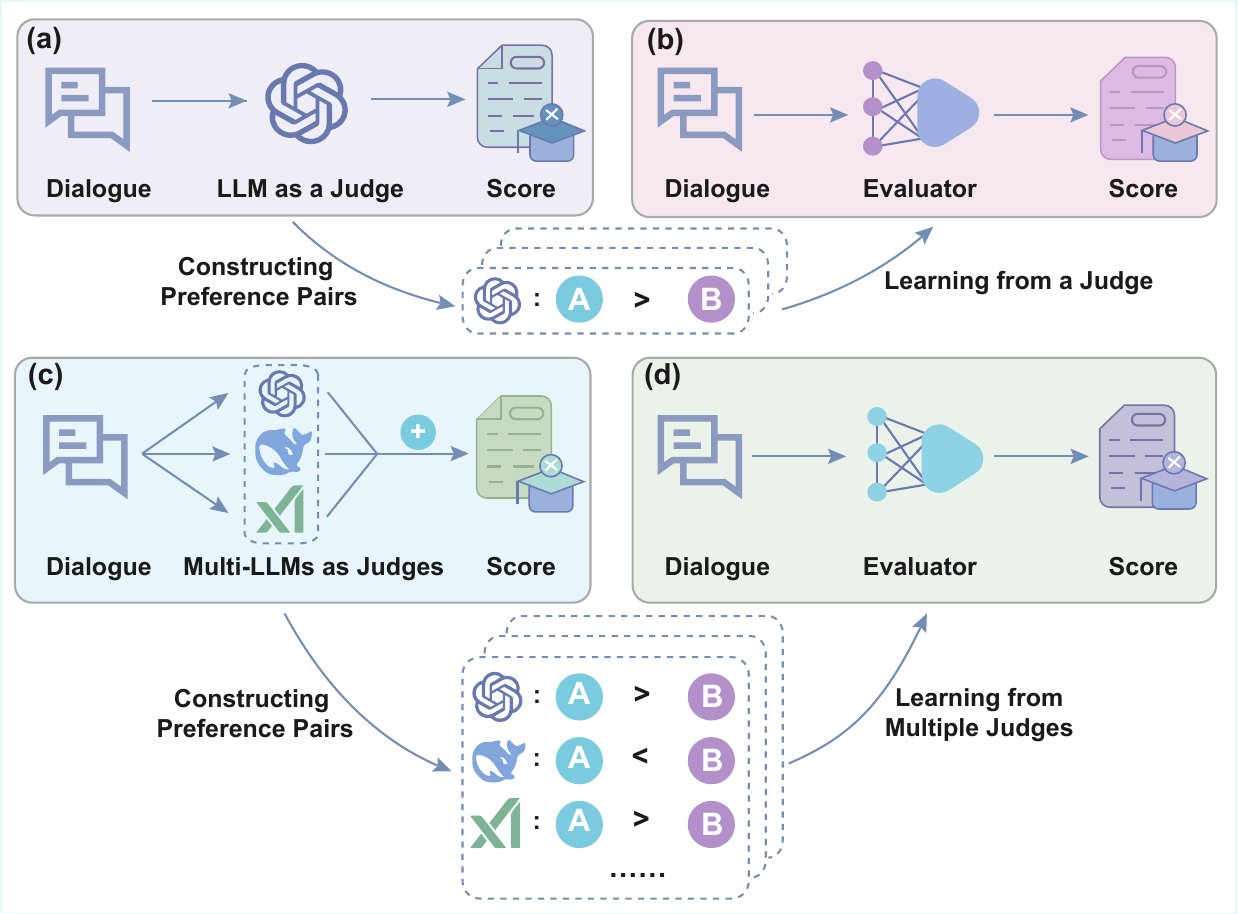}
  \caption{Comparison among different dialogue evaluation paradigms. (a) The traditional LLM-as-a-Judge approach. (b) Learning an evaluator using the preference data from a judge. (c) The aggregation of multiple LLMs for evaluation. (d) Learning an evaluator from multiple LLM judges (Ours).}
  \label{fig:Motivation_Method_Compare}
  \vspace{-1em}
\end{figure}

In recent years, the “LLM-as-a-judge” paradigm \cite{zheng2023judging,gu2024survey,li2024generation,li2025exploring} has gained increasing attention for assessing response quality through single rating or pairwise comparison (Fig.~\ref{fig:Motivation_Method_Compare}). While proprietary LLM judges show high correlation with human judgments \cite{zheng2023judging,liu2023g,zhou2023lima}, they often suffer from opaque training data, limited controllability, and high inference cost \cite{kim2024prometheus}. To improve transparency and deployability, recent works have explored training open-source LLM evaluators, typically by learning from preference data provided by a single judge \cite{sahoo2025quantitative,li2023generative,zhu2023judgelm}. However, these learned evaluators still lag behind in robustness and scalability, and single-judge annotation is known to exhibit systematic biases such as self-preference, score compression, and prompt sensitivity \cite{thakur2024judging,wei2024systematic}. To mitigate these biases, recent methods \cite{rahmani2024judgeblender,verga2024replacing,sun2024skillaggregation} employ multiple LLM judges and aggregate their outputs to obtain more reliable assessments. Although effective, these multi-judge approaches incur substantial inference-time overhead, which hinders scalability and real-time use in large-scale dialogue evaluation.

To address these challenges, we propose an efficient multi-turn dialogue evaluator that learns from multiple LLM judges. 
Given that pairwise comparisons are more reliable than direct scoring for LLMs \cite{cui2023ultrafeedback,kim2024prometheus,feng2025samer}, we first construct a large-scale pairwise preference dataset for multi-turn dialogues, with each dialogue annotated by multiple advanced LLM judges. We then develop a lightweight evaluator composed of a text-embedding model with specialized scoring heads, using a learning-to-rank training strategy \cite{liu2009learning}. In particular, we explicitly incorporate a maximum likelihood estimation approach to jointly optimize the evaluator and model the reliability of each judge. Extensive experiments on various multi-turn dialogue benchmarks show that our model exhibits strong robustness and wide applicability, surpassing existing baselines across various evaluation tasks. Our main contributions are as follows:
\begin{itemize}
    \item We propose a learning framework that effectively aggregates preference data from multiple LLM judges into a single evaluator. This reference-free method retains the strengths of diverse multi-judge feedback while substantially reducing computation costs.
    \item We develop \textbf{\OURS{}}, a lightweight open-source model that can efficiently and flexibly evaluate multi-turn dialogues for both single rating and pairwise comparison tasks. \OURS{} supports evaluations at overall or fine-grained levels across ten commonly used dialogue quality dimensions.
    \item We construct \textbf{\OURDATA{}}, a large-scale preference dataset for multi-turn dialogues, where each response pair is annotated by multiple advanced LLM judges across ten fine-grained dimensions as well as an overall quality, providing rich supervision for training evaluators.
    \item We release \textbf{Daily-MTD}, a high-quality human-annotated benchmark comprising 600 multi-turn dialogues focused on daily-life scenarios. Each dialogue is annotated with overall ratings, pairwise preferences, and fine-grained judgments across ten evaluation dimensions.
\end{itemize}

\section{Related Work}

\subsection{LLM-based Dialogue Evaluation}
In the past two years, LLM-as-a-judge \cite{zheng2023judging,gu2024survey,li2023alpacaeval,chan2023chateval,thakur2024judging} has gained attention as a promising paradigm to mimic the depth and granularity of human in multi-turn dialogue evaluation \cite{kim2024prometheus}. Inspired by the high correlations with human evaluation capabilities of advanced proprietary LLMs,
recent research \cite{wang2023pandalm,zeng2023evaluating,cui2023ultrafeedback,lee2024prometheus} has emphasized the development of open-source LLM evaluators. 
Prometheus2 \cite{kim2024prometheus} enhances adaptability to custom evaluation criteria via weight merging, supporting both direct assessment and pairwise ranking tasks. SaMer \cite{feng2025samer} introduces a scenario-aware, multi-dimensional evaluation framework that dynamically selects relevant evaluation dimensions based on the specific dialogue scenario, thereby enabling both fine-grained and overall assessment. 
However, these models primarily focus on single-turn dialogue evaluation. Although they can be extended to assess multi-turn dialogues, their performance in this setting is notably degraded. We aim to fill this gap by introducing \OURS{}, a dialogue evaluator specifically trained for multi-turn dialogue quality assessment.

\subsection{Multi-LLM Aggregation for Evaluation}
A range of aggregation methods \cite{rahmani2024judgeblender,sun2024skillaggregation,wei2024measuring,sun2024crosscheckgpt} have been widely explored for LLM-based evaluation. 
Simple methods such as majority voting \cite{bai2024mt}, averaging probabilities \cite{sun2024skillaggregation}, and Crowdlayer \cite{rodrigues2018deep} have demonstrated the effectiveness of leveraging multiple LLMs to obtain robust results. 
Recent studies, such as PoLL \cite{verga2024replacing}, utilize a straightforward aggregation approach by assigning equal weight to each evaluator. Similarly, PRD \cite{li2023prd} employs multiple LLMs for pairwise ranking. 
Despite their effectiveness, these methods entail substantial computational overhead due to the need for repeated inference across multiple models, which limits their applicability in large-scale or real-time scenarios. In this study, we aim to develop an efficient evaluator by learning from multiple LLM judges.


\begin{figure*}[t]
  \centering
  \includegraphics[width=\textwidth]{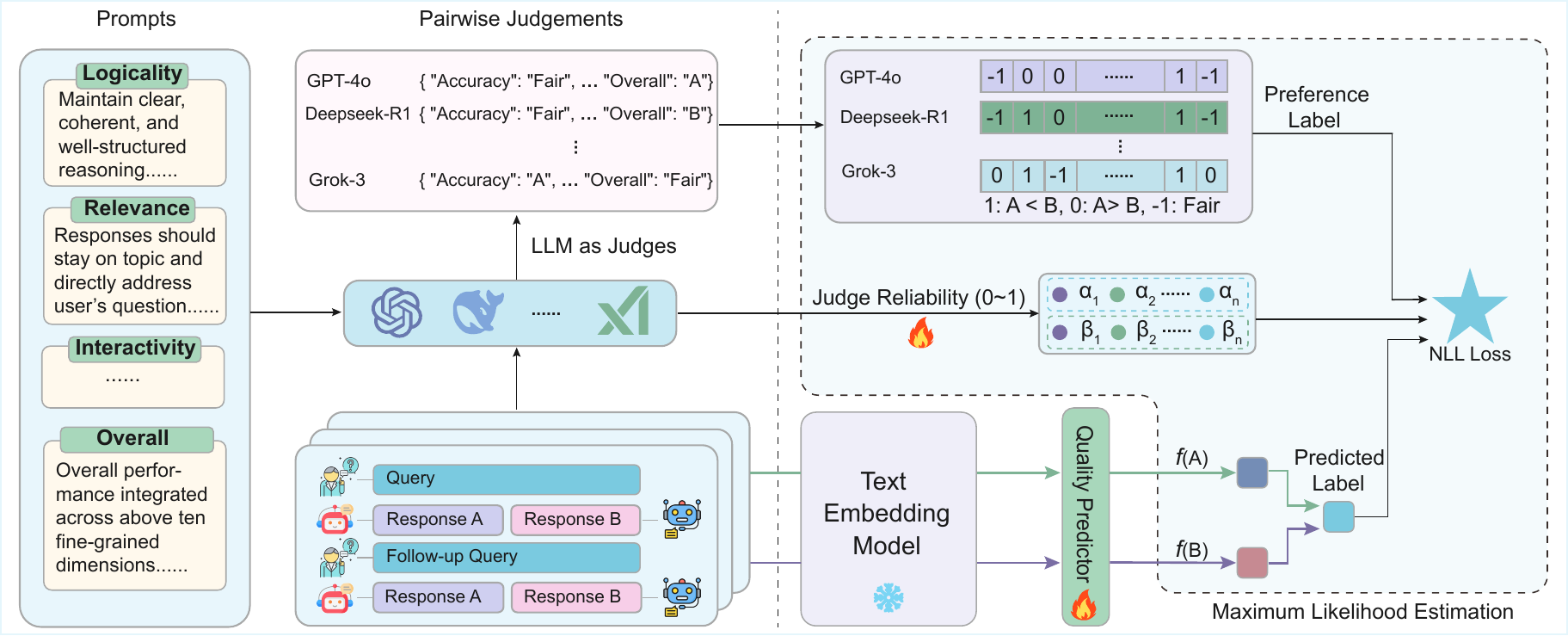}
  \caption{Overview of \OURS{}. The left part illustrates the construction of a multi-judge-annotated, fine-grained preference dataset used for training. The right part demonstrates the model architecture and training procedure, which comprises an LLM-based text embedding model and an MLP-based quality prediction head. The training involves a probabilistic formulation of pairwise preferences with judge reliability prediction, which is optimized by maximum likelihood estimation.}
  \label{fig:proposed_method}
  \vspace{-1em}
\end{figure*}

\section{Dataset Construction}\label{sec:data_construct}

\subsection{Training Dataset: \OURDATA{}} \label{sec:training_data_annotate}
Since existing pairwise preference datasets focus on single-turn dialogues and typically annotated by a single judge \cite{alex_havrilla_2023,kim2024prometheus,ji2024pku}, in this study, we construct \textbf{\OURDATA{}}, a novel \textbf{\underline{P}}airwise \textbf{\underline{P}}reference-annotated \textbf{\underline{M}}ulti-\textbf{\underline{T}}urn \textbf{\underline{D}}ialogue dataset, where each dialogue pair is evaluated across {ten fine-grained evaluation dimensions} plus an overall rating by multiple LLM judges. 

Specifically, we construct \OURDATA{} based on the \textit{Multiturn-Chat-0.8M} dataset \cite{wen2023chathome}, which contains 831K Chinese multi-turn dialogues. We employ five advanced LLMs as judges, namely Claude-3.7-Sonnet \cite{anthropic2024claude}, GPT-4o \cite{openai2024gpt4ocard}, Grok-3 \cite{grok3beta}, DeepSeek-R1 \cite{deepseekai2025deepseekr1incentivizingreasoningcapability}, and Gemini-2.0-Flash \cite{team2023gemini}. Each judge is guided by a standardized evaluation template, which requires them to first provide a detailed explanation, followed by an evaluation along ten fine-grained dimensions, \ie, Accuracy, Logicality, Conversationality, Relevance, Personalization, Creativity, Interactivity, Emotionality, Informativeness, and Safety. Each dimension, along with the overall score, is rated on a three-alternative forced choice: ``A'', ``B'', or ``Fair''. The data pre-processing procedures are provided in Appendix \ref{appendix_a_train_data}.

Finally, we construct a multi-judge annotated preference dataset comprising $\sim$11K multi-turn dialogues. Each instance consists of two dialogues responding to the same user query, along with preference annotations $\{r^{j}\}_{j=1}^M$ from $M = 5$ LLM judges. Each judge $j$ provides evaluations $r^j = \{r^j_1, r^j_2, \ldots, r^j_K, r^j_\text{all}\}$ where $K = 10$ represents fine-grained evaluation dimensions and $r^j_\text{all}$ denotes the overall preference. This dataset is then used for training a dialogue quality evaluator.

\subsection{Evaluation Benchmark: Daily-MTD}\label{sec:human_annotated}
To provide a reliable benchmark for evaluating multi-turn dialogue quality, we introduce a novel high-quality human-annotated evaluation dataset named \textbf{Daily-MTD}, consisting of 600 carefully curated multi-turn dialogues in daily-life scenarios. Specifically, we recruited ten undergraduate students majoring in Computer Science to participate in dialogue collection tasks. Each participant interacted with two distinct chatbots, engaging in conversations ranging from 2 to 10 turns. This process yielded an initial pool of 1,080 multi-turn dialogues. Subsequently, five NLP researchers independently selected 600 diverse dialogues from the initial collection. Each expert annotates the dialogues across 10 fine-grained dimensions, along with an overall quality rating. 
The final labels for single rating were obtained by score averaging, while those for pairwise and multi-dimensional comparisons were determined via majority voting. Finally, we constructed a high-quality human-annotated evaluation dataset, adapted into three specialized forms to accommodate different assessment tasks: Daily-MTD for single rating, Daily-MTD-Pair for pairwise comparison, and Daily-MTD-Dim for multi-dimensional comparison.
More details about Daily-MTD are presented in Appendix \ref{ap:Detail_Daily_MTD}.

\section{Proposed Method}
In this section, we present \OURS{}, an efficient \textbf{\underline{M}}ulti-\textbf{\underline{T}}urn \textbf{\underline{D}}ialogue \textbf{\underline{Eval}}uator learned from multiple judges. As illustrated in Figure~\ref{fig:proposed_method}, \OURS{} involves: (1) a probabilistic formulation of pairwise preferences with \emph{dimension-aware} judge reliability estimation, and (2) parameter optimization via Maximum Likelihood Estimation.

\subsection{Problem Formulation}
Let $\mathcal{D} = \{(A_i, B_i), \mathbf{r}_i^1, \ldots, \mathbf{r}_i^M\}_{i=1}^{N}$ be a dataset of $N$ multi-turn dialogue pairs, where each pair $(A_i, B_i)$ is annotated by $M$ LLM judges over $K$ fine-grained evaluation dimensions \emph{and} an overall criterion. We use $k \in \{1,\ldots,K,\text{all}\}$ to index these dimensions. Each judge $j$ outputs preference vector $\mathbf{r}^j \coloneqq \mathbf{r}^j(A,B)=\{r_k^j(A,B)\}_{k\in\{1,\ldots,K,\text{all}\}}$, where each $r_k^j$ takes one of three outcomes:

\begin{itemize}
  \item $ r_k^j(A, B) = 0 $. The judge $j$ prefers $A$ over $B$ on dimension $k$.
  \item $ r_k^j(A, B) = 1 $. Conversely, $B$ is preferred over $A$ on dimension $k$.
  \item $ r_k^j(A, B) = -1 $. $A$ and $B$ are considered equally preferable (\emph{Fair}) on dimension $k$.
\end{itemize}

By leveraging pairwise signals from multiple LLM judges, our goal is to learn an efficient evaluator on $\mathcal{D}$ that estimates dialogue quality. \OURS{} consists of a \textit{frozen} embedding model $f_g$ parameterized by $g$ and a \textit{trainable} quality predictor $f_\omega$ with parameters $\omega$ that comprises \textbf{two multi-layer perceptron (MLP) heads}: (i) a $K$-dimensional head producing fine-grained scores $\{f_{\omega \circ g,k}(x)\}_{k=1}^{K}$, and (ii) an overall head producing $f_{\omega \circ g,\text{all}}(x)$. Both heads are trained under the same formulation below, conditioning on the corresponding $k$. The construction of $\mathcal{D}$ is described in Section~\ref{sec:training_data_annotate}.

\subsection{Probabilistic Modeling}
Let $q_k(x)$ denote the (latent) quality of a multi-turn dialogue $x$ on dimension $k$, and $z_k \in \{0,1\}$ denote the latent ground-truth preference label on dimension $k$ for a dialogue pair $(A,B)$. To model uncertainty in dialogue quality, we adopt \textit{Thurstone's Case V model} \cite{thurstone2017law}. Specifically, we assume that $q_k(x)$ follows a Gaussian distribution with mean $f_{\omega \circ g,k}(x)$ and standard deviation $\sigma$, \ie, $q_k(x) \sim \mathcal{N}(f_{\omega \circ g,k}(x), \sigma)$. Under this assumption, the quality difference $q_k(A)-q_k(B)$ follows $\mathcal{N}\big(f_{\omega \circ g,k}(A) - f_{\omega \circ g,k}(B),\, 2\sigma^2\big)$. The probability that dialogue $B$ has higher quality than $A$ (\ie, the probability of $ z_k = 1 $) is then
\begin{equation}\label{eq:1}
\resizebox{0.89\linewidth}{!}{$
\begin{aligned}
\Pr\bigl(z_k=1|A,B,g,\omega\bigr)
&= \Pr\bigl(q_k(A)<q_k(B)\mid g, \omega \bigr) \\
&= \Phi\!\left(\tfrac{f_{\omega \circ g,k}(B)-f_{\omega \circ g,k}(A)}{\sqrt{2}\sigma}\right),
\end{aligned}
$}
\end{equation}
where $\Phi(\cdot)$ is the standard normal cumulative distribution function (CDF).

To quantify the reliabilities of LLM judges, we assume that they can be characterized by probabilities of correct annotations (known as ``hits" rate and ``correct rejection" rate in signal detection theory \cite{ma2019blind}) for the input dialogue pair \(A\) and \(B\). If the latent ground-truth label on dimension $k$ is $z_k=1$ (favoring B), the hit rate of judge $j$ is
\begin{equation}
\alpha_k^{j} = \Pr(r_k^{j} = 1 \mid z_k = 1).
\label{eq:alpha}
\end{equation}
Similarly, when $z_k=0$, the correct rejection rate is
\begin{equation}
\beta_k^{j} = \Pr(r_k^{j} = 0 \mid z_k = 0).
\label{eq:beta}
\end{equation}
The reliability parameters $\{\alpha_k^j,\beta_k^j\}$ can be jointly estimated with the trainable parameters $\omega$. It is worth noting that "Fair" cases are deliberately excluded, which means that when $r_k^{j}=-1$, neither $\alpha_k^j$ nor $\beta_k^j$ is updated.

\subsection{Maximum Likelihood Estimation}
We formulate the parameter optimization as a maximum likelihood estimation task. Given the assumption that judgments across training dialogue pairs are independent, we can factorize the likelihood function over the full set of parameters $\{\omega, \alpha, \beta \}$ as
\begin{equation}\label{eq:4}
\resizebox{1\linewidth}{!}{$
\Pr\big(\mathcal{D}\mid \omega, \alpha, \beta\big)
= \prod_{i=1}^{N}\prod_{k=1}^{K} \Pr(r_{ik}^1, \ldots, r_{ik}^{M_{ik}^{'}}\mid A_i, B_i; g, \omega, \alpha_k, \beta_k),
$}
\end{equation}
where $M_{ik}^{'}$ refers to the number of judges with non-\emph{Fair} labels for $(A_i,B_i)$ on dimension $k$. Since $r_{ik}^j$ is conditionally independent given the judge-specific reliability parameters $\{\alpha_k^j,\beta_k^j\}$ and the ground-truth preference label $z_{ik}$, we decompose the likelihood by conditioning on $z_{ik}$
\begin{equation}\label{eq:5}
\resizebox{0.89\linewidth}{!}{$
\begin{aligned}
  & \Pr\big(r_{ik}^1, \ldots, r_{ik}^{M_{ik}^{'}}\mid A_i, B_i; g, \omega, \alpha_k, \beta_k\big) =
   \\
  & \Pr(r_{ik}^1, \ldots, r_{ik}^{M_{ik}^{'}}\mid z_{ik} = 1, \alpha_k)\Pr(z_{ik} = 1 \mid A_i, B_i; g, \omega) \ + \\
  & \Pr(r_{ik}^1, \ldots, r_{ik}^{M_{ik}^{'}}\mid z_{ik} = 0, \beta_k)\Pr(z_{ik} = 0 \mid A_i, B_i; g, \omega).
\end{aligned}
$}
\end{equation}
The distribution modeling the noisy annotations from LLM judges can be expressed as:
\begin{equation}\label{eq:6}
\resizebox{0.89\linewidth}{!}{$
\begin{aligned}
  \Pr\big(r_{ik}^1, \ldots, r_{ik}^{M_{ik}^{'}}\mid z_{ik} = 1, &\alpha_k\big)
  = \prod_{j=1}^{M_{ik}^{'}} \Pr(r_{ik}^j\mid z_{ik} = 1, \alpha_k^j) \\
  &= \prod_{j=1}^{M_{ik}^{'}} (\alpha_k^j)^{r_{ik}^j} (1-\alpha_k^j)^{1-r_{ik}^j},
\end{aligned}
$}
\end{equation}
and
\begin{equation}\label{eq:7}
\resizebox{0.89\linewidth}{!}{$
\begin{aligned}
  \Pr\big(r_{ik}^1, \ldots, r_{ik}^{M_{ik}^{'}}\mid z_{ik} = 0, \beta_k\big)
  = \prod_{j=1}^{M_{ik}^{'}} (\beta_k^j)^{1-r_{ik}^j} (1-\beta_k^j)^{r_{ik}^j}.
\end{aligned}
$}
\end{equation}
Denoting the probability $\Pr(z_{ik}=1\mid A_i,B_i;g,\omega)$ from Eq. (\ref{eq:1}) as $\mathcal{P}_k(A_i,B_i;g,\omega)$, and the reliability terms from Eqs. (\ref{eq:6}) and (\ref{eq:7}) as $\mathcal{A}_k(z_{ik},\alpha_k)$ and $\mathcal{B}_k(z_{ik},\beta_k)$ respectively, and substituting them into Eq. (\ref{eq:4}), we obtain the complete likelihood function
\begin{equation}\label{eq:8}
\resizebox{0.89\linewidth}{!}{$
\begin{aligned}
  \Pr\big(\mathcal{D}\mid \omega, \alpha, \beta\big)
  &= \prod_{i=1}^{N}\prod_{k=1}^{K}
  \Big{[}\mathcal{A}_k(z_{ik}, \alpha_k)\mathcal{P}_k(A_i, B_i; g, \omega) \\
  &+\ \mathcal{B}_k(z_{ik}, \beta_k)\big(1 - \mathcal{P}_k(A_i, B_i; g, \omega)\big)\Big{]}.
\end{aligned}
$}
\end{equation}
We minimize the Negative Log-Likelihood (NLL)
\begin{equation}
  \ell(\omega, \alpha, \beta) = -\log\Pr\big(\mathcal{D}\mid \omega, \alpha, \beta\big),
  \label{eq:loss}
\end{equation}
using stochastic gradient descent to obtain the optimal parameters $\{\hat{\omega}, \hat{\alpha}, \hat{\beta}\}$ for \OURS{}.

\section{Experiments}
In this section, we demonstrate the effectiveness of \OURS{} across comprehensive dialogue evaluation tasks. We first describe our implementation setup (Section~\ref{subsec:training_details}) and evaluation datasets and metrics covering single rating, pairwise comparison, and multi-dimensional comparison tasks (Section~\ref{subsec:benchmarks}). We then present the main results of \OURS{}, showing significant improvements over open-source baselines and competitive performance against proprietary LLMs (Section~\ref{subsec:main_result}). Finally, we conduct ablation studies on the effects of judge composition (Section~\ref{subsec:ablation}), analyze the learned reliability parameters of different judges (Section~\ref{subsec:judge_reliable}), and compare inference efficiency against baseline approaches (Section~\ref{subsec:inference}).

\subsection{Implementation of \OURS{}}\label{subsec:training_details}

We employ the Llama-3-8B as the text embedding model $g$ and initialize it with parameters from a trained Llama-3-8B reward model~\citep{wang2024interpretable}. An MLP layer $\omega$, serving as a quality predictor, is appended to the embedding model. This MLP is optimized using the loss defined in Eq.~(\ref{eq:loss}), while keeping the embedding model frozen.
During training, we initially set $\sigma=1$ in Eq. (\ref{eq:1}), and $\alpha = \beta = 0.5$ in Eq. (\ref{eq:alpha}) and (\ref{eq:beta}). The model is trained for 3 epochs with a batch size of 32. More training details are presented in Appendix \ref{ap:Training_Details}.

\begin{table*}[!t]
\centering
\renewcommand{\arraystretch}{1.1}
\setlength{\tabcolsep}{1.1mm}
\vspace{-0.5em}
\resizebox{1.0\textwidth}{!}{
\begin{tabular}{lccccccccccccccc}
\toprule
\multirow{3}{*}{\textbf{Evaluator}} & \multicolumn{6}{c}{\textbf{Single Rating}} & \multicolumn{8}{c}{\textbf{Pairwise Comparison}}\\
\cmidrule(lr){2-7}\cmidrule(lr){8-15} & \multicolumn{2}{c}{xDial-IEval} & \multicolumn{2}{c}{MT-Bench} & \multicolumn{2}{c}{Daily-MTD} & \multicolumn{2}{c}{xDial-IEval-Pair} & \multicolumn{2}{c}{MT-Bench-Human} & \multicolumn{2}{c}{Chatbot-Arena} & \multicolumn{2}{c}{Daily-MTD-Pair} \\
\cmidrule(lr){2-3}
\cmidrule(lr){4-5}
\cmidrule(lr){6-7}
\cmidrule(lr){8-9}
\cmidrule(lr){10-11}
\cmidrule(lr){12-13}
\cmidrule(lr){14-15}
& Pearson & Spearman & Pearson & Spearman & Pearson & Spearman & w/ TIE & w/o TIE & w/ TIE & w/o TIE & w/ TIE & w/o TIE & w/ TIE & w/o TIE \\
\midrule
GPT-4o                   & 0.615 & 0.641 & 0.708 & 0.707 & 0.691 & 0.440 & 80.51 & 89.28 & 77.30 & 90.58 & 68.63 & 79.53 & 64.86 & 78.18 \\
Grok-3                   & 0.598 & 0.612 & 0.673 & 0.657 & 0.687 & 0.448 & 79.92 & 88.41 & 80.26 & 91.93 & 69.72 & 79.92 & 66.67 & 79.41 \\
Claude-3.7-Sonnet        & 0.599 & 0.606 & 0.686 & 0.698 & 0.665 & 0.460 & 80.94 & 90.14 & 79.93 & 91.58 & 68.93 & 80.22 & 66.33 & 79.69 \\
Deepseek-R1              & 0.606 & 0.617 & 0.630 & 0.649 & 0.669 & 0.396 & 81.27 & 89.57 & 80.92 & 91.23 & 70.16 & 81.17 & 68.00 & 80.67 \\
Gemini-3-Pro  & 0.588 & 0.620 & 0.675 & 0.681 & 0.714 & 0.475 & 78.18 & 90.28 & 81.97 & 92.63 & 71.17 & 81.95 & 67.42 & 81.46 \\
GPT-5.1  & 0.605 & 0.608 & 0.676 & 0.699 & 0.684 & 0.398 & 79.98 & 89.05 & 81.52 & 91.49 & 69.92 & 79.57 & 66.57 & 79.92 \\
\midrule
Majority Vote  & 0.575 & 0.585 & 0.710 & 0.736 & 0.671 & 0.390 & 82.17 & 90.72 & 82.09 & 93.84 & 71.97 & 82.28 & 71.67 & 82.85 \\
Average Scoring  & 0.617 & 0.677 & 0.739 & 0.762 & 0.689 & 0.392 & -- & -- & -- & -- & -- & -- & -- & -- \\
\midrule
Llama-3.1-8B-Instruct    & 0.438 & 0.441 & 0.442 & 0.440 & 0.613 & 0.341 & 70.13 & \textbf{86.96} & 59.21 & 78.60 & 51.93 & 65.60 & 55.50 & 69.33 \\
Qwen2.5-7B-Instruct      & 0.594 & 0.592 & 0.550 & 0.514 & 0.620 & 0.396 & 63.54 & 80.58 & 54.93 & 79.56 & 53.40 & 66.96 & 58.33 & 69.33 \\
Qwen3-8B  & \underline{0.599} & \underline{0.615} & \textbf{0.628} & \textbf{0.630} & \underline{0.654} & \underline{0.402} & \underline{74.95} & 85.67 & \underline{71.72} & \underline{85.42} & \underline{62.06} & \underline{76.50} & \underline{64.33} & \underline{79.77} \\
DynaEval-125M  & 0.368 & 0.372 & -0.109 & -0.132 & 0.134 & 0.101 & 58.73 & 68.41 & 36.18 & 55.09 & 43.28 & 50.58 & 43.33 & 54.20 \\
AutoJ-13B                & 0.540 & 0.541 & 0.432 & 0.408 & 0.629 & 0.379 & 49.12 & 62.37 & 49.11 & 76.70 & 52.72 & 62.88 & 53.97 & 67.20 \\
Prometheus-7B            & 0.322 & 0.317 & 0.378 & 0.385 & 0.336 & 0.194 & 42.28 & 59.14 & 41.54 & 52.54 & 41.13 & 45.91 & 36.67 & 42.71 \\
Prometheus-2-7B          & 0.591 & 0.607 & 0.548 & 0.509 & 0.594 & 0.357 & 67.30 & 82.30 & 50.45 & 73.48 & 55.85 & 59.94 & 62.20 & 71.15 \\
ArmoRM-8B                & 0.519 & 0.547 & 0.455 & 0.427 & 0.589 & 0.373 & 69.87 & 84.64 & 64.80 & 80.53 & 57.80 & 68.95 & 61.33 & 76.37 \\
SaMer-8B                    & 0.434 & 0.422 & 0.384 & 0.377 & 0.496 & 0.372 & 59.75 & 75.52 & 58.86 & 73.94 & 61.56 & 72.46 & 53.11 & 68.45 \\
\midrule
\textbf{\OURS{} (8B)}  & \textbf{0.612} & \textbf{0.617} & \underline{0.587} & \underline{0.554} &  \textbf{0.662} & {\textbf{0.523}} & \textbf{76.70} & \underline{86.10} & \textbf{79.01} & \textbf{91.93} & {\textbf{71.24}} & \textbf{81.29} & \textbf{65.33} & {\textbf{81.93}} \\
\bottomrule
\end{tabular}
}
\caption{Evaluation results on three {single rating} and four {pairwise comparison} benchmarks. The best result among non-SOTA LLMs is bolded, and the second best is underlined.}
\label{tab:main_result}
\vspace{-0.6em}
\end{table*}

\subsection{Evaluation Datasets and Metrics}\label{subsec:benchmarks}
We conduct a comprehensive evaluation of \OURS{} using three types of benchmarks: overall-level single rating and pairwise comparison, as well as multi-dimensional comparison. Detailed descriptions of the evaluation datasets are provided in Appendix \ref{ap:eval_bench}.

\begin{itemize}
\item \textbf{Single Rating}: This paradigm evaluates the response independently, enabling absolute quality assessment without reliance on comparative baselines. We measure the correlation between model-generated scores and human reference ratings using Pearson and Spearman coefficients. Evaluations are performed on three benchmarks: xDial-IEval~\cite{svikhnushina2022ieval}, MT-Bench~\cite{zheng2023judging}, and Daily-MTD.

\item \textbf{Pairwise Comparison}: This protocol assesses an evaluator’s ability to predict human preferences between two competing responses. We adopt two evaluation modes: one excluding tie cases (\emph{w/o tie}) and another including them (\emph{w/ tie}), to evaluate both binary and ternary judgment accuracy. The evaluation is conducted on four benchmarks: xDial-IEval-Pair~\cite{svikhnushina2022ieval}, MT-Bench-Human~\cite{zheng2023judging}, Chatbot-Arena~\cite{chiang2024chatbot}, and Daily-MTD-Pair.

\item \textbf{Multi-Dimensional Comparison} requires evaluators to generate independent preference judgments for response pairs based on a specific evaluation dimension. The evaluation is performed on Daily-MTD-Dim. This benchmark allows us to validate whether evaluators can capture nuanced, aspect-level differences in multi-turn dialogues.
\end{itemize}

\subsection{Baselines}\label{subsec:baselines}
To evaluate the performance of \OURS{}, we compare against a diverse set of LLM-based baselines, categorized into the following four groups. (1) \textbf{State-of-the-art LLMs}, including GPT-4o~\cite{openai2024gpt4ocard}, Grok-3~\cite{grok3beta}, Claude-3.7-Sonnet~\cite{anthropic2024claude}, DeepSeek-R1~\cite{deepseekai2025deepseekr1incentivizingreasoningcapability}, Gemini-3-Pro~\cite{team2023gemini}, and GPT-5.1~\cite{openai2025gpt5systemcard}. These models serve as strong upper-bound references in the LLM-as-a-judge paradigm. 
(2) \textbf{Classical aggregation methods}, including Majority Vote and Average Scoring, which aggregate predictions from above SOTA LLMs to establish a consensus-based reference.
(3) \textbf{Open-source LLMs}, including Llama-3.1-8B Instruct ~\cite{dubey2024llama}, Qwen2.5-7B-Instruct~\cite {qwen2}, and Qwen3-8B ~\cite{yang2025qwen3}, which represent mid-sized models with comparable scale to our evaluator.  
(4) \textbf{Specialized LLM Evaluator}, including DynaEval-125M~\cite{zhang2021dynaeval}, AutoJ-13B~\cite{li2023generative}, SaMer-8B~\cite{feng2025samer}, Prometheus-7B~\cite{kim2023prometheus}, Prometheus2-7B~\cite{kim2024prometheus}, and ArmoRM-8B~\cite{wang2024interpretable}, all explicitly trained for dialogue quality assessment.
LLMBar~\cite{zeng2023evaluating}. 

\subsection{Main Results}\label{subsec:main_result}
We first compare \OURS{} with baselines using three benchmark settings: single rating, pairwise comparison, and multi-dimensional comparison.

\begin{table*}[!ht]
\centering
\renewcommand{\arraystretch}{1.15}
\setlength{\tabcolsep}{4mm}
\resizebox{\textwidth}{!}{
\begin{tabular}{lccccccccccccc}
\toprule
\textbf{Evaluator} & Acc. & Log. & Conv. & Rel. & Pers. & Crea. & Inter. & Emo. & Info. & Safe. & Average \\
\midrule
GPT-4o                   & 85.67 & 79.67 & 58.67 & 85.67 & 63.00 & 67.67 & 59.00 & 54.00 & 67.67 & 92.00 & 71.30 \\
Grok-3                   & 75.33 & 72.00 & 57.33 & 74.67 & 51.33 & 56.00 & 55.67 & 50.33 & 63.67 & 92.33 & 64.87 \\
Claude-3.7-Sonnet        & 76.33 & 69.00 & 63.33 & 73.00 & 56.33 & 60.33 & 60.00 & 58.33 & 67.00 & 91.33 & 67.50 \\
Deepseek-R1              & 76.67 & 77.73 & 61.00 & 80.67 & 59.67 & 64.00 & 62.00 & 57.67 & 70.00 & 97.00 & 70.64 \\  
Gemini-3-Pro  & 90.67 & 82.33 & 59.00 & 85.67 & 68.67 & 69.00 & 78.00 & 64.67 & 63.33 & 92.67 & 75.40 \\
GPT-5.1  & 84.00 & 72.33 & 51.00 & 83.00 & 69.67 & 59.33 & 64.33 & 62.33 & 65.00 & 83.33 & 69.43  \\
\midrule
Majority Vote  & 86.96 & 86.29 & 53.56 & 89.90 & 55.36 & 62.27 & 72.89 & 68.76 & 66.19 & 92.68 & 73.49  \\
\midrule
Llama-3.1-8B-Instruct    & 43.67 & 52.33 & 60.33 & 52.67 & 71.67 & 48.00 & 45.00 & 60.00 & 37.00 & 64.67 & 53.53 \\
Qwen2.5-7B-Instruct      & \underline{76.67} & \underline{72.33} & 68.67 & 77.67 & \underline{74.00} & 61.00 & \textbf{61.33} & 64.00 & 63.33 & 64.67 & 68.37 \\
Qwen3-8B  & \textbf{79.59} & 70.82 & \underline{72.37} & \textbf{85.98} & 68.97 & \underline{65.36} & \underline{59.79} & \underline{66.19} & \underline{72.97} & \underline{81.08} & \underline{72.31}  \\
AutoJ-13B                & 34.67 & 29.67 & 43.00 & 39.67 & 62.67 & 55.67 & 35.33 & 57.00 & 33.67 & 22.33 & 41.37 \\
Prometheus-7B            & 21.14 & 24.07 & 33.44 & 26.50 & 44.03 & 32.44 & 29.76 & 42.03 & 22.44 & 10.41 & 28.63 \\
Prometheus-2-7B          & 68.33 & \textbf{73.67} & 69.92 & \underline{82.33} & 64.00 & 48.67 & 58.72 & 66.00 & 63.67 & 70.33 & 66.56 \\
\midrule
\textbf{\OURS{} (8B)}    & 72.33 & 68.33 & \boxed{\textbf{79.00}} & 67.67 & \boxed{\textbf{75.67}} & \boxed{\textbf{74.67}} & 52.33 & \boxed{\textbf{66.33}} & \boxed{\textbf{74.67}} & \boxed{\textbf{97.67}} & \textbf{72.87} \\
\bottomrule
\end{tabular}
}
\caption{Multi-dimensional evaluation results on the Daily-MTD-Dim dataset. The best and second-best results among non-SOTA LLMs are bolded and underlined, respectively. Boxes highlight cases where our method outperforms all SOTA LLMs.}
\label{tab:Fine_Grained_Dim}
\vspace{-0.5em}
\end{table*}

\begin{figure*}[!t]
  \centering
  \includegraphics[width=\textwidth]{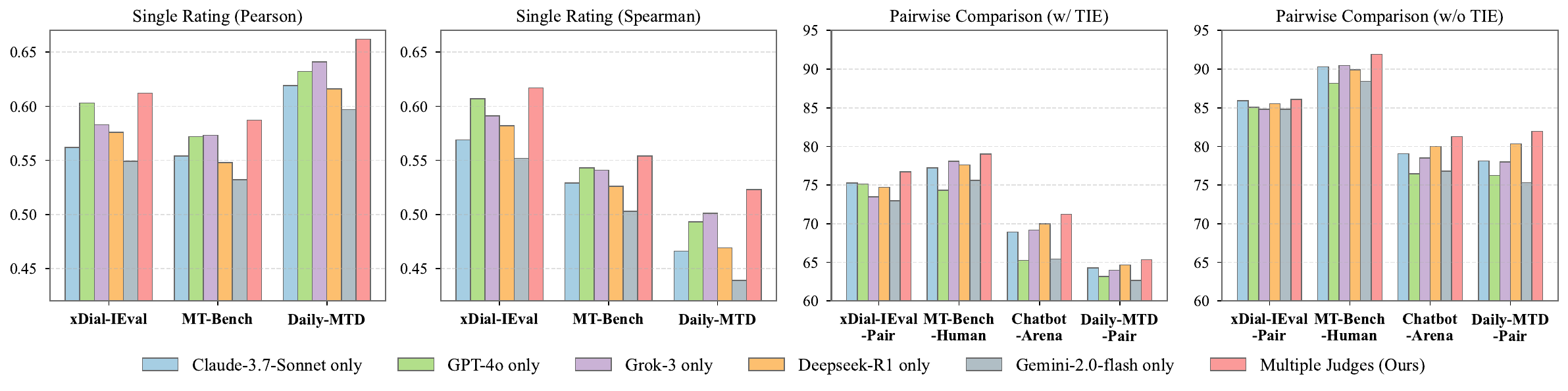}
  \caption{Performance comparison of models trained on annotations from individual LLM judges versus our model trained on multi-judge preferences.}
  \label{fig:ablation_study_I}
  \vspace{-0.6em}
\end{figure*}

\begin{table*}[h]
\centering
\renewcommand{\arraystretch}{1.2}
\setlength{\tabcolsep}{0.5mm}
\resizebox{1.0\textwidth}{!}{
\begin{tabular}{lcccccccccccccccc}
\toprule
\multirow{3}{*}{\textbf{Evaluator}} 
    & \multicolumn{6}{c}{\textbf{Single Rating}} 
    & \multicolumn{8}{c}{\textbf{Pairwise Comparison}} \\
\cmidrule(lr){2-7} \cmidrule(lr){8-15}
    & \multicolumn{2}{c}{xDial-IEval} 
    & \multicolumn{2}{c}{MT-Bench} 
    & \multicolumn{2}{c}{Daily-MTD} 
    & \multicolumn{2}{c}{xDial-IEval-Pair} 
    & \multicolumn{2}{c}{MT-Bench-Human} 
    & \multicolumn{2}{c}{Chatbot-Arena} 
    & \multicolumn{2}{c}{Daily-MTD-Pair} \\
\cmidrule(lr){2-3} \cmidrule(lr){4-5} \cmidrule(lr){6-7} 
\cmidrule(lr){8-9} \cmidrule(lr){10-11} \cmidrule(lr){12-13} \cmidrule(lr){14-15}
    & Pearson & Spearman & Pearson & Spearman & Pearson & Spearman
    & w/ TIE & w/o TIE & w/ TIE & w/o TIE & w/ TIE & w/o TIE & w/ TIE & w/o TIE \\
\midrule
\textbf{Default} (5 Judges)  & 0.612 & 0.617 & 0.587 & 0.554 & 0.662 & 0.523 & 76.70 & 86.10 & 79.01 & 91.93 & 71.24 & 81.29 & 65.33 & 81.93 \\
\midrule
- Deepseek-R1   & -0.006 & -0.008 & -0.008 & -0.012 & -0.019 & -0.016 & -1.33 & -0.30 & -1.59 & -0.86 & -0.59 & -0.77 & -0.33 & -0.94 \\
+ gpt-4o-mini   & -0.004 & -0.005 & -0.002 & -0.003 & -0.006 & -0.009 & -0.50 & -0.59 & -0.71 & 0.00 & -0.27 & -0.55 & -1.00 & -0.82 \\
+ o4-mini       & +0.005 & +0.006 & +0.009 & +0.005 & +0.005 & +0.002 & +0.52 & +0.20 & +2.24 & +0.35 & +1.07 & -0.40 & +0.50 & +0.42 \\
\bottomrule
\end{tabular}
}
\caption{Performance of \OURS{} under varying compositions of LLM judges.}
\label{tab:ablation_study_II}
\vspace{-0.6em}
\end{table*}

In \textbf{single rating} tasks (Table~\ref{tab:main_result}), \OURS{} demonstrates significant improvements across all three benchmarks, surpassing open-source baselines by considerable margins. The performance gains are particularly notable on xDial-IEval, where \OURS{} even outperforms most proprietary models. 
Furthermore, \OURS{} achieves correlation coefficient improvements exceeding 10\% compared to its ArmoRM backbone across all benchmarks, providing robust evidence of our framework's effectiveness. 
However, all open-source LLMs still underperform relative to advanced proprietary LLMs on MT-Bench, underscoring the persistent challenges posed by this benchmark and highlighting the existing performance gap with proprietary models.

In \textbf{pairwise comparison} tasks (Table~\ref{tab:main_result}), \OURS{} establishes clear superiority among open-source LLMs, securing top performance in 7 out of 8 tasks across four benchmarks and achieving second place in the remaining task. Although ArmoRM-8B already demonstrates competitive performance as an open-source baseline, \OURS{} achieves at least 5\% improvements across most tasks over this strong foundation. These gains reach approximately 15\% on particularly challenging benchmarks, such as MT-Bench-Human and Chatbot Arena. Notably, \OURS{} outperforms almost all proprietary LLMs on several pairwise comparison datasets. This provides compelling evidence of its exceptional cross-task generalization capabilities.

In \textbf{multi-dimensional comparison} tasks (Table~\ref{tab:Fine_Grained_Dim}), results show that dimension-level accuracy drops significantly for most evaluators, underscoring the inherent challenge of precise assessment on the specific dimension. Importantly, \OURS{} achieves substantial improvements over existing open-source models across most dimensions, \eg, Conversationality and Safety. In particular, \OURS{} delivers a 31.5\% improvement compared to its original Llama-3-8B-Instruct backbone, and surpasses most proprietary LLMs in averaged dimension-level accuracy. These findings collectively validate the efficacy of \OURS{}'s training framework and establish its superior performance in multi-dimensional dialogue evaluation.

\subsection{Ablation Study}\label{subsec:ablation}
\paragraph{Single Judge vs. Multiple Judges} To further analyze the effectiveness of our method, we conduct ablation studies on \OURS{} by examining the impact of leveraging multiple LLM judges during training. As shown in Figure~\ref{fig:ablation_study_I}, we compare \OURS{} trained on preference annotations from all five LLM judges against variants trained using annotations from each individual judge only.
The results show that the method of learning from multiple judges consistently surpasses all of them across both single rating and pairwise comparison benchmarks. This demonstrates that our framework successfully integrates complementary signals from multiple LLM judges, effectively distilling their collective expertise into a more robust and generalizable evaluator.

\paragraph{Judge Quality and Quantity} We also examine how judge {quality} and {quantity} influence model performance by varying judge composition. We evaluate several configurations: (i) removing the most reliable judge (DeepSeek-R1), (ii) replacing a high-performing judge with a weaker one (GPT-4o-mini), and (iii) augmenting the judges with an additional strong judge (o4-mini). Results are summarized in Table~\ref{tab:ablation_study_II}. The findings show that incorporating higher-quality judges consistently improves evaluator performance, although the gains are modest. In contrast, removing reliable judges or introducing lower-capability ones leads to measurable performance degradation. These results indicate that both the quantity and, more importantly, the quality of LLM judges play crucial roles in shaping the effectiveness of the learned evaluator within our framework.


\begin{figure}[t]
  \centering
  \includegraphics[width=0.9\linewidth]{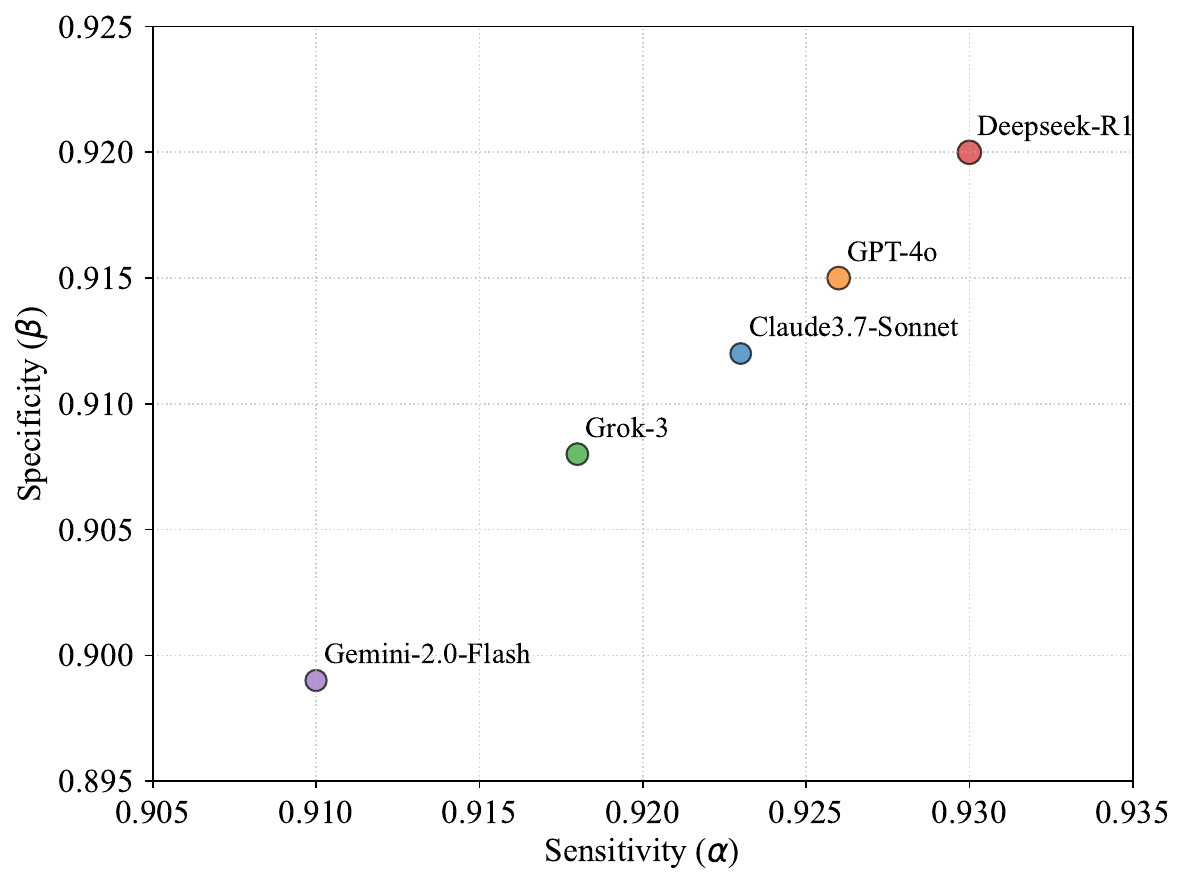}
  \caption{The learned sensitivity ($\alpha$) and specificity ($\beta$) of the five LLM judges.}
  \label{fig:alpha_beta_value}
  \vspace{-0.6em}
\end{figure}

\subsection{Judge Reliability Analysis}\label{subsec:judge_reliable}
The learned sensitivity ($\alpha$) and specificity ($\beta$) values for the five LLM judges are shown in Figure \ref{fig:alpha_beta_value}. As expected, our method preferentially weights judges with stronger predictive capabilities, such that those exhibiting higher accuracy in dialogue quality assessment exert greater influence during the model's learning process. Furthermore, the learned $\alpha$ and $\beta$ parameters exhibit a strong correlation, which suggests their potential substitutability with a unified parameter vector during training.

\begin{table}[t] 
\centering
\renewcommand{\arraystretch}{1.1}
\setlength{\tabcolsep}{1.5mm} 
\resizebox{1.0\linewidth}{!}{
\begin{tabular}{lcc}
\toprule
\textbf{Evaluator} & Single Rating & Pairwise Comparison \\
\midrule
Llama-3.1-8B-Instruct    & 0.57 & 1.11   \\
Qwen2.5-7B-Instruct      & 0.46 & 0.87   \\
Qwen3-8B-Instruct        & 1.33 & 2.33   \\
DynaEval-125M                 & 0.13 & 0.24   \\ 
AutoJ-13B                & 2.32 & 4.77   \\
Prometheus-7B            & 0.69 & 1.79   \\
Prometheus-2-7B          & 1.36 & 3.23   \\
ArmoRM-8B                & 0.75 & 1.53   \\
SaMer-8B                 & 0.23 & 0.45   \\
\midrule
\textbf{\OURS{} (8B)}    & \textbf{0.10} & \textbf{0.19}  \\
\bottomrule
\end{tabular}
}
\caption{Inference efficiency comparison of \OURS{} against baselines on Daily-MTD and Daily-MTD-Pair (average runtime in seconds per instance).}
\label{tab:Inference_Efficiency}
\vspace{-0.6em}
\end{table}

\subsection{Inference Efficiency Comparison}\label{subsec:inference}
We further compare the inference efficiency of \OURS{} with other open-source LLMs of similar parameter scale (7$\sim$8B parameters) on the Daily-MTD dataset. All experiments are conducted on dual NVIDIA GeForce RTX 4090 GPUs with models loaded in bf16 precision, using the HuggingFace Transformers library. As shown in Table~\ref{tab:Inference_Efficiency}, \OURS{} achieves substantially lower average runtime than baseline models, demonstrating superior inference efficiency. This advantage is primarily due to \OURS{}'s encoder-based architecture, which outputs quality scores in a single forward pass through MLP layers. In contrast, autoregressive LLM baselines rely on generative evaluation, resulting in longer latency and higher computational cost.

\section{Conclusion}

In this work, we present a learning framework for multi-turn dialogue evaluation that aggregates preference knowledge from multiple LLM judges into a single evaluator. Built upon this framework, we develop {\OURS{}}, an efficient evaluator capable of flexible assessment of dialogue quality through both single rating and pairwise comparison. Moreover, we contribute two high-quality resources to the community: {\OURDATA{}}, a large-scale multi-judge annotated preference dataset for training dialogue evaluators, and {Daily-MTD}, a human-annotated evaluation benchmark with fine-grained annotations across ten dialogue quality dimensions. Extensive experiments have demonstrated that \OURS{} outperforms existing baselines in terms of accuracy, robustness, and inference efficiency.

\section*{Limitations}
Despite these advances, several limitations remain. 
First, our reliance on multiple SOTA LLMs as judges for fine-grained preferences inevitably incurs substantial annotation costs, which may reduce the feasibility of scaling the training dataset to larger scales.
Additionally, current training data is primarily focused on daily-life scenarios, potentially limiting generalization to more specialized domains. For future work, we plan to expand the coverage of evaluation scenarios and multilingual dialogues, and develop a more nuanced evaluation framework with cost-efficient supervision to further enhance the generalization of learned evaluators.



\bibliography{ref}

\clearpage

\appendix

\section*{Appendix}

\section{Details of \OURDATA{} Construction}\label{appendix_a_train_data}
\subsection{Dataset Preprocessing}
The detailed preprocessing procedures for \OURDATA{} are as follows:
\begin{itemize}
    \item We filtered the data to retain multi-turn dialogues consisting of 2 to 10 turns.
    \item The dialogue direction is naturally controlled by the user without interference from the assistant, and the content is free from excessive technical terminology or irrelevant information through keyword-based selection; for example, dialogue turns containing only trivial utterances such as ``Hello,'' ``Sorry,'' or ``Thank you'' were excluded.
    \item Leveraging DeepSeek-V3 \cite{deepseekai2024deepseekv3technicalreport}, we automatically filtered dialogues that maintain thematic consistency, logical coherence, and smooth transitions between responses, and then translated them into English.
    \item Different models (gpt-4o-mini \cite{openai2024gpt4ocard}, Qwen2-72B-Instruct \cite{qwen2}, gemini-1.5-flash \cite{team2024gemini}, and claude-3.5-sonnet \cite{anthropic2024claude}) were used to generate alternative responses with noticeable quality differences. To mitigate length bias, we further ensured that the difference in response lengths between each response pair did not exceed 10 words.
\end{itemize}



\subsection{Data Annotation}\label{appendix_a2_data_annotation}
The following details outline the specific steps taken during data annotation:
\begin{itemize}
    \item To prevent potential overfitting and self-preference, where an LLM evaluator rates its own output higher than others’ while human judges would consider them of equal quality \cite{panickssery2024llm}, we specifically select five SOTA LLMs distinct from those used during the data generation process as judges.
    \item To eliminate position bias, we swapped the A/B order of each dialogue pair, retaining only those samples where all five judges provided consistent outputs after the swap.
    \item To ensure a balanced distribution of annotation labels, the final dataset was further filtered to contain 40\% “A”, 40\% “B” and 20\% “Fair” annotations.
\end{itemize}

The following prompt was used to annotate the training data for multi-turn dialogues, both fine-grained and overall.
\begin{tcolorbox}[colback=blue!3, colframe=black, arc=2mm, left=3pt, right=3pt,
    boxsep=5pt, boxrule=0.5pt, colframe=black!70, fonttitle=\bfseries, title = Data Annotation Prompt, breakable]

\textbf{System Prompt} \\
You are an expert in evaluating multi-turn dialogues between AI assistants and users. 
Your task is to compare two dialogues across 10 fine-grained dimensions and an overall evaluation. For each dimension, choose exactly one of: "A", "B", or "Fair". Respond strictly in the specified JSON format. \\

\textbf{User Prompt} \\
\#\#\# Task Description \#\#\# \\
Evaluate two multi-turn dialogues between a user and AI assistants across 10 fine-grained dimensions and an overall dimension. For each dimension: \\
- If Assistant A performs better, choose "A" \\
- If Assistant B performs better, choose "B" \\
- If performance is equivalent or indistinguishable, choose "Fair" \\

\#\#\# Evaluation Criteria \#\#\# \\
  \textbf{Accuracy}: Ensure factual consistency and reliability throughout the dialogue. Avoid hallucinations, contradictions, or misleading content. Responses should interpret questions accurately and provide appropriate support. \\
  \textbf{Logicality}: Maintain clear, coherent, and well-structured reasoning. Avoid contradictions or confusion, and ensure causal chains are preserved across turns.  \\
  \textbf{Conversationality}: Language should be fluent, natural, and human-like. Avoid mechanical or awkward phrasing. Use ellipsis, pronouns, and context-sensitive expressions where appropriate. \\
  \textbf{Relevance}: Responses should stay on topic and directly address the user's question. Avoid digressions and ensure context is appropriately inherited across turns.  \\
  \textbf{Personalization}: Responses should exhibit a unique and expressive style. If user preferences are explicit, adapt to them. Avoid overfitting to user style.  \\
  \textbf{Creativity}: Provide novel, engaging, and thoughtful responses beyond template-based or memorized content. Encourage interesting interactions.  \\
  \textbf{Interactivity}: Manage the flow of dialogue proactively. Balance information density with cognitive load and encourage continued conversation when appropriate. \\
  \textbf{Emotionality}: Recognize and adapt to emotional cues in conversation. Express emotions appropriately (e.g., humor, empathy) to maintain warm and natural dialogue.  \\
  \textbf{Informativeness}: Dynamically assess the user's need for depth. Provide verifiable, valuable knowledge rather than shallow or generic answers. \\
  \textbf{Safety}: Continuously monitor risks across the dialogue. Avoid suggestive traps, follow ethical norms, and prevent misleading, offensive, or unsafe responses. \\
  \textbf{Overall}: Overall performance integrated across above ten fine-grained dimensions. \\

\#\#\# Output Format (STRICT) \#\#\# \\
Respond with exactly 11 lines of JSON objects: \\
\{"Accuracy": "A/B/Fair"\} \\
\{"Logicality": "A/B/Fair"\} \\
\{"Conversationality": "A/B/Fair"\} \\
\{"Relevance": "A/B/Fair"\} \\
\{"Personalization": "A/B/Fair"\} \\
\{"Creativity": "A/B/Fair"\} \\
\{"Interactivity": "A/B/Fair"\} \\
\{"Emotionality": "A/B/Fair"\} \\
\{"Informativeness": "A/B/Fair"\} \\
\{"Safety": "A/B/Fair"\} \\
\{"Overall": "A/B/Fair"\} \\

\#\#\# Dialogue A \#\#\# \\
\{dialogue\_A\} \\

\#\#\# Dialogue B \#\#\# \\
\{dialogue\_B\} \\

Now begin your structured response:

\end{tcolorbox}

Below is an annotated example of our training data.

\begin{tcolorbox}[colback=blue!3, colframe=black, arc=2mm, left=3pt,right=3pt,
    boxsep=5pt, boxrule=0.5pt, colframe=black!70, fonttitle=\bfseries, title = Training Data Example, breakable]

\vspace{3pt}
\noindent\textbf{Conversation:}
\begin{itemize}
  \item \textbf{Human:} Can you create a poem about summer?
    \begin{itemize}
      \item \textbf{Assistant A:} Beneath the sun’s warm, golden gaze, The world awakens in a vibrant haze. Fields of green and skies of blue, Whisper secrets, fresh and new. Laughter dances on the breeze, As summer weaves its sweetest ease. A fleeting joy, yet ever true, This season’s gift, for me and you.
      \item \textbf{Assistant B:} The summer sunlight is warm and vibrant, with flowers and grass swaying gently in the breeze. The morning chorus of birds and the evening serenade of cicadas resonate throughout the air, complementing the exquisite atmosphere of summer. The freshness and beauty of the season remain etched in our hearts.
    \end{itemize}

  \item \textbf{Human:} Can you help me think of some outdoor activities that are suitable for summer?
    \begin{itemize}
      \item \textbf{Assistant A:} Swimming in pools or lakes, hiking scenic trails, having backyard barbecues, and playing beach volleyball are perfect summer activities. You could also try camping, cycling through parks, or organizing picnics with friends and family to enjoy the warm weather.
      \item \textbf{Assistant B:} Certainly. Summer outdoor activities are exceptionally diverse, including swimming, picnicking, camping, hiking, cycling, kayaking, beach volleyball, and more. These activities not only allow individuals to enjoy the pleasures of outdoor sports but also provide opportunities to spend quality time with friends and family.
    \end{itemize}

  \item \textbf{Human:} Can you tell me some healthy foods for the summer?
    \begin{itemize}
      \item \textbf{Assistant A:} Fresh fruits like watermelon, berries, and peaches are hydrating and refreshing. Salads with leafy greens, cucumbers, and tomatoes are light yet nutritious. Grilled fish, smoothies, and chilled soups like gazpacho also make excellent, healthy summer meal options.
      \item \textbf{Assistant B:} Certainly. Healthy foods for the summer include fruits such as watermelon, strawberries, and blueberries, as well as cold vegetables and salads, delicious seafood, and grilled meats. These foods are rich in essential nutrients and contribute positively to maintaining overall health.
    \end{itemize}
\end{itemize}

\vspace{3pt}
\noindent\textbf{Evaluations:}
\begin{itemize}
  \item[] \textbf{Claude\texttt{-}3.7\texttt{-}sonnet:}  \{
     Accuracy: Fair, Logicality: B, Conversationality: A, Relevance: Fair, 
     Personalization: A, Creativity: A, Interactivity: Fair, Emotionality: A,
     Informativeness: Fair, Safety: Fair, Overall: A \}
  
  \item[] \textbf{GPT\texttt{-}4o:} \{
     Accuracy: Fair, Logicality: Fair, Conversationality: Fair, Relevance: Fair,
     Personalization: Fair, Creativity: A, Interactivity: Fair, Emotionality: Fair,
     Informativeness: Fair, Safety: Fair, Overall: A \}
  
  \item[] \textbf{Grok\texttt{-}3:} \{
     Accuracy: Fair, Logicality: Fair, Conversationality: A, Relevance: Fair, 
     Personalization: A, Creativity: A, Interactivity: Fair, Emotionality: A,
     Informativeness: Fair, Safety: Fair, Overall: A \}
  
  \item[] \textbf{Deepseek\texttt{-}R1:} \{
     Accuracy: Fair, Logicality: Fair, Conversationality: A, Relevance: Fair,
     Personalization: A, Creativity: A, Interactivity: Fair, Emotionality: Fair, 
     Informativeness: Fair, Safety: Fair, Overall: A \}
  
  \item[] \textbf{Gemini\texttt{-}2.0\texttt{-}flash:} \{
     Accuracy: Fair, Logicality: Fair, Conversationality: A, Relevance: Fair, 
     Personalization: Fair, Creativity: A, Interactivity: Fair, Emotionality: Fair, 
     Informativeness: Fair, Safety: Fair, Overall: A \}
\end{itemize}

\end{tcolorbox}

\begin{table}[h] 
\centering
\renewcommand{\arraystretch}{1.2}
\setlength{\tabcolsep}{1.5mm} 
\resizebox{0.8\linewidth}{!}{
\begin{tabular}{lccc}
\toprule
\multirow{2}{*}{\textbf{Judge}} & \multicolumn{3}{c}{\textbf{Preference Labels}} \\ 
\cmidrule(lr){2-4}
 & A & B & Fair \\ 
\midrule
Claude-3.7-Sonnet & 4,652 & 4,777 & 1,902 \\ 
GPT-4o & 4,013 & 4,530 & 2,788 \\
Grok-3 & 4,623 & 4,578 & 2,130 \\
DeepSeek-R1 & 4,910 & 4,834 & 1,587  \\
Gemini-2.0-Flash & 4,580 & 4,525 & 2,226  \\
\bottomrule
\end{tabular}
}
\caption{Statistics of judges' preference}
\label{tab:Training_Data_Annotation}
\end{table}


\section{Details of Daily-MTD Construction}\label{ap:Detail_Daily_MTD}
Ten undergraduate students majoring in Computer Science were recruited to participate in dialogue data collection tasks. Each participant engaged in interactions with two distinct chatbots, generating dialogues consisting of 2 to 10 conversational turns. Participants received a \$10 gift card upon completing every 15 dialogue tasks, with each student eligible to contribute up to 120 dialogues. Dialogue tasks were administered through personalized links, each directing the participants to a designated set of tasks hosted on an internal chat interface deployed on a local server. This procedure resulted in an initial corpus of 1,080 multi-turn dialogues.

Subsequently, five experts in the field of NLP independently curated a representative set of 600 diverse dialogues from the initial collection. Each expert performed annotations across 10 fine-grained dimensions, accompanied by an overall quality rating, using the same annotation prompt detailed in Appendix~\ref{appendix_a2_data_annotation}. Annotators received compensation at a rate of \$5 per 10 dialogues annotated, totaling \$750. The annotation results demonstrated substantial inter-annotator agreement; specifically, in terms of overall quality ratings, 45\% of the dialogues received unanimous ratings from all five annotators, 25\% had agreement from four annotators, and 30\% from three annotators. Final labels for single-rating evaluations were determined via score averaging, whereas labels for pairwise and multi-dimensional comparisons were established through majority voting. Ultimately, we developed a high-quality, human-annotated evaluation dataset, structured into three specialized subsets catering to distinct evaluation tasks: Daily-MTD for single-rating tasks, Daily-MTD-Pair for pairwise comparisons, and Daily-MTD-Dim for multi-dimensional assessments.


\section{Details of \OURS{} Training} \label{ap:Training_Details}
The optimization of \OURS{} is facilitated by the DeepSpeed library~\citep{rasley2020deepspeed}, leveraging the ZeRO Stage 2 optimizer~\citep{rajbhandari2020zero} and FlashAttention2~\citep{dao2023flashattention} to achieve efficient parallelism across two NVIDIA GeForce RTX 4090 GPUs. We utilize the AdamW optimizer~\citep{loshchilov2017decoupled}, configured with $\beta_1 = 0.9$, $\beta_2 = 0.95$, and a weight decay of 0.1. The primary learning rate for the model is set to $5 \times 10^{-5}$, whereas the learning rates for $\alpha$ and $\beta$ are specified as $1 \times 10^{-2}$. A linear warm-up is applied during the initial 10\% of training steps, followed by a cosine decay schedule to zero. We employ a batch size of 32 and limit the maximum sequence length to 8,192 tokens. The model is trained for 3 epochs to ensure convergence and optimal performance.

The proposed method supports both overall and fine-grained scoring. Accordingly, we trained two distinct models in our experiments: one for overall rating and another for evaluating the performance across specific dimensions.


\section{Introduction of Evaluation Datasets}\label{ap:eval_bench}
We conduct a comprehensive evaluation of \OURS{} using three types of benchmarks: overall-level single rating and pairwise comparison, as well as multi-dimensional comparison.

\textbf{Single Rating} is regarded as an effective evaluation strategy as it dispenses with the need for comparative baselines. However, its reliance on the LLM's intrinsic judgment may introduce subjectivity and overlook nuanced distinctions. In single rating, we use Pearson and Spearman as performance metrics to measure scoring correlations with reference answers using a reference-based method. The three Single Rating benchmarks are:

\begin{itemize}
    \item \textbf{xDial-IEval} \cite{svikhnushina2022ieval}: A multi-turn dialogue evaluation benchmark consisting of 1,920 dialogue instances and 8 score rubrics, with each dialogue averaging 6 turns, rated by human evaluators on a scale from 1 to 5.
    
    \item \textbf{MT-Bench} \cite{zheng2023judging}: A multi-turn chat benchmark that consists of 80 curated prompts, 80 expert-defined scoring rubrics \cite{kim2023prometheus}; and 320 model responses generated by WizardLM-13B, Vicuna-13B, LLaMA-2-13B-Chat, and GPT-3.5-Turbo-0613. 

    \item \textbf{Daily-MTD}: A meticulously constructed multi-turn daily dialogue benchmark consisting of 600 instances, 10 fine-grained score rubrics, each independently annotated by five experts specializing in NLP. The evaluation is conducted on a scale from 1 to 10, and the final score for each instance is derived through a score averaging method across all judges.
\end{itemize}
\textbf{Pairwise Comparison} reduces subjectivity while enabling the identification of finer distinctions by comparing model-generated response pairs. In pairwise comparison, we use two evaluation approaches: first, excluding "tie" cases (denoted as ‘w/o tie’); second, grouping "tie" responses for evaluation (denoted as ‘w/ tie’). And we use a reference-free method to measure the alignment between model judgments and human judgments. The four Pairwise Comparison benchmarks are:
\begin{itemize}
    \item \textbf{xDial-IEval-Pair} 
    \cite{svikhnushina2022ieval}: A multi-turn benchmark that leverages the original xDial-IEval dataset. For each set of four identical user queries, two response pairs are randomly selected and labeled as ‘win’, ‘fair’, or ‘lose’ based on their respective scores, resulting in the creation of 480 pairwise comparison instances.

    \item \textbf{MT-Bench-Human} \cite{zheng2023judging}: A multi-turn benchmark that utilizes the same 80 test prompts as MT-Bench. Furthermore, it provides 3,360 response pairs, which are judged by human evaluators as ‘win’, ‘tie’, or ‘lose’. 
    
    \item \textbf{Chatbot-Arena} \cite{chiang2024chatbot}: A benchmark that consists of 30K arena data, from which we randomly sample 2K multi-turn dialogues. The dataset includes responses from models such as GPT-4, GPT-3.5, Claude, Vicuna-7B/13B, Koala-13B, LLaMA-13B. All judgments are provided by collected crowd judges.

    \item \textbf{Daily-MTD-Pair}: A benchmark that utilizes the original Daily-MTD dataset. For each user query, two response pairs A/B are available, and based on their respective scores, they are labeled as "A", "Fair", or "B", which leads to the generation of 300 pairs.
\end{itemize}

\textbf{Multi-Dimensional Comparison} requires evaluators to generate independent preference judgments for response pairs based on specific dimensions. To assess this capability, we constructed the Daily-MTD-Dim dataset using the annotation information for two alternative responses to each user query in the original Daily-MTD dataset across ten dimensions, ultimately yielding 300 pairs.

\newcolumntype{P}[1]{>{\raggedright\arraybackslash}p{#1}}
\begin{table*}[!htbp]
  \small
  \renewcommand{\arraystretch}{1.3}
  \resizebox{\textwidth}{!}{%
    \begin{tabular}{P{4.2cm} P{2.4cm} P{1.8cm} P{1.8cm} P{6cm}}
      \toprule
      \textbf{Model Name} & \textbf{Creator} & \textbf{\#Parameters} & \textbf{Access}       & \textbf{URL}                 \\
      \midrule
      GPT-4o-2024-11-20  & OpenAI  & undisclosed  & Official API  & \texttt{https://chat.openai.com}   \\
      Grok-3-2025-02-17  & xAI  & undisclosed & Official API  & \texttt{https://x.ai}   \\
      Claude-3.7-Sonnet-2025-02-19 & Anthropic & undisclosed & Official API & \texttt{https://claude.ai}         \\
      DeepSeek-R1-2025-01-20  & DeepSeek  & 671B & Official API & \texttt{https://www.deepseek.com}  \\
       Gemini-3-Pro-2025-11-18  & Google Deepmind & unclosed & official API & \texttt{https://gemini.google.com}  \\
       GPT-5.1-2025-11-13  & OpenAI & unclosed & official API & \texttt{https://chat.openai.com}    \\
      \midrule
      Llama3.1-8B-Instruct & Meta & 8B & Weights & \texttt{https://llama.meta.com/llama3} \\
      Qwen2.5-7B-Instruct & Alibaba & 7B & Weights & \texttt{https://qwenlm.github.io/} \\
      Qwen3-8B  & Alibaba & 8B & Weights & \texttt{https://qwenlm.github.io/}  \\
      \midrule
      DynaEval-125M & NUS & 125M & Weights & \texttt{https://github.com/e0397123/DynaEval} \\
      AutoJ-13B  & Gair & 13B & Weights & \texttt{https://huggingface.co/GAIR} \\
      Prometheus-7B  & prometheus-eval & 7B & Weights & \texttt{https://huggingface.co/prometheus-eval} \\
      Prometheus-2-7B  & prometheus-eval & 7B  & Weights & \texttt{https://huggingface.co/prometheus-eval} \\
      ArmoRM-8B  & RLHFlow & 8B  & Weights  & \texttt{https://huggingface.co/RLHFlow}        \\
       SaMer-8B  & ZJU-HIC & 8B  & Weights & \texttt{https://github.com/Irving-Feng/SaMer}        \\
      \bottomrule
    \end{tabular}%
  }
  \caption{Overview of the baselines evaluated in our experiments.}
  \label{tab:baseline_description}
\end{table*}


\section{Introduction of Baseline Models}\label{ap:models}
We have selected 15 high-performing LLMs as our baselines. GPT-4o-2024-11-20, Grok-3-2025-02-17, Claude-3.7-Sonnet-2025-02-19, Deepseek-R1-2025-01-20, Gemini-3-Pro-2025-11-18, GPT-5.1-2025-11-13 are accessed via their official APIs. The remaining open-source models are deployed on a local server equipped with two NVIDIA GeForce RTX 4090 GPUs. To ensure fair comparisons, we evaluated all models using their original prompt templates and manually remove the reference answer module from the prompt templates of the Prometheus series. For reward models such as ArmoRM-8B and SaMer-8B, we directly use their original architectures to score the dialogues. The detailed information of these models is shown in Table \ref{tab:baseline_description}.

\section{Ethical Statement}

This study aims to advance the reliability and efficiency of automatic evaluation methods for multi-turn dialogue systems. The data used in \OURS{} primarily target everyday-life scenarios, and the supervision signals are mainly derived from preference judgments generated by multiple SOTA LLMs, together with controlled annotations provided by human experts. This design ensures that no privacy or data confidentiality risks are involved.

We emphasize that \OURS{} is strictly a research-oriented evaluation model intended for assessing multi-turn dialogues and is not designed for deployment in real-world, high-risk settings. Given its reliance on synthetic supervision and simulated evaluation environments, this study is not expected to raise significant ethical concerns.

\section{Detailed Prompts for Evaluation}\label{appendix_c_evaluation}
To ensure consistency and reproducibility for multi-turn dialogue evaluation, we design standardized prompts for both single rating and pairwise comparison tasks, requiring all LLM judges to strictly follow the specified response formats. Below are the default prompt templates for these two evaluation tasks.

\begin{tcolorbox}[colback=blue!3, colframe=black, arc=2mm, left=3pt,right=3pt,
    boxsep=5pt, boxrule=0.5pt, colframe=black!70, fonttitle=\bfseries, 
    title = Prompts for Single Rating Evaluation, breakable]

\textbf{System Prompt} \\[2pt]
You are an expert in evaluating the quality of dialogues. \\
Your task is to assess the performance and quality of an AI assistant in multi-turn conversations.

\vspace{6pt}
\textbf{User Prompt} \\[2pt]
\#\#\# Evaluation Guidelines \#\#\#
\begin{enumerate}
  \item Rate the assistant's performance on a scale from 1 to 10, where a higher score indicates better quality.
  \item Consider the following dimensions when assigning the overall rating: Accuracy, Logicality, Conversationality, Relevance, Personalization, Creativity, Interactivity, Emotionality, Informativeness, Safety.
  \item Criteria for rating:
  \begin{itemize}
    \item 1–3 points: Performance is poor, with most or all criteria not being met or significantly lacking.
    \item 4–6 points: Performance is average, with some aspects meeting the standards, but overall there are notable deficiencies.
    \item 7–9 points: Performance is good, with most aspects meeting high standards.
    \item 10 points: Performance is exceptional, with all criteria being excellently met.
  \end{itemize}
  \item Avoid any potential bias. Your judgment should be as objective as possible.
\end{enumerate} 

Here is the dialogue you need to rate: \\
\{dialogue\} \\

\vspace{6pt}
\#\#\# Instructions for Rating \#\#\#
\begin{enumerate}
  \item Make your judgment based on the evaluation guidelines.
  \item Provide the final score — an integer between 1 and 10.
  \item Begin with the following response formats: \\
  \{Rating: [[X]]\} \\
   where X is the score you assign.
\end{enumerate}

\end{tcolorbox}

\begin{tcolorbox}[colback=blue!3, colframe=black, arc=2mm, left=3pt,right=3pt,
    boxsep=5pt, boxrule=0.5pt, colframe=black!70, fonttitle=\bfseries, title = Prompts for Pairwise Comparison Evaluation, breakable]

\textbf{System Prompt} \\
You are an expert in evaluating the quality of dialogues. \\
Your task is to objectively and impartially evaluate the overall performance and quality of two AI assistants throughout their multi-turn dialogue.\\
\#\#\# Evaluation rules: \#\#\# \\
If Assistant A performs better than Assistant B, respond with "A"; \\
if Assistant B performs better than Assistant A, respond with "B"; \\
If the performance of Assistant A and Assistant B is equivalent, indistinguishable, respond with "Fair". \\

\textbf{User Prompt} \\
Here are two dialogues you need to compare:\\
dialogue A: \\
\{dialogue\_A\} \\

dialogue B: \\
\{dialogue\_B\} \\ \\
\#\#\# Below are the instructions to assess and compare the overall quality of two responses: \#\#\# \\   
1. Make your judgment based on the evaluation rules \\
2. Begin your response with one of these exact formats: \\
So my final decision is: \{\{"Overall": "A"\}\} or \{\{"Overall": "B"\}\} or \{\{"Overall": "Fair"\}\}  \\
        
\end{tcolorbox}


\section{Examples of Evaluation Results} \label{appendix_d_case_study}
In this part, we demonstrate three detailed examples regarding three types of benchmarks to illustrate the performance of different models, thereby highlighting the superior capability of our model across various multi-turn dialogue evaluation tasks.
\begin{tcolorbox}[colback=gray!3, colframe=black, arc=2mm, left=3pt,right=3pt,
    boxsep=5pt, boxrule=0.5pt, colframe=black!70, fonttitle=\bfseries, title = A Case of Single Rating on Daily-MTD Dataset, breakable]

\vspace{3pt}
\noindent\textbf{Conversation:}
\begin{itemize}

  \item \textbf{Human:} Hi, can I ask you a question?
    \begin{itemize}
      \item \textbf{Assistant:} Questions are often asked, and sometimes they can be interesting or not.
    \end{itemize}

  \item \textbf{Human:} I’d like to know how to prevent my computer from getting viruses.
    \begin{itemize}
      \item \textbf{Assistant:} To prevent viruses, ensure your computer is running, and maybe check the lights. Sometimes, turning it off helps, but that might not be related to viruses.
    \end{itemize}

  \item \textbf{Human:} Sure, thanks for your suggestion!
    \begin{itemize}
      \item \textbf{Assistant:} Suggestions are helpful sometimes, but other times they don't really matter. It depends on the situation.
    \end{itemize}
\end{itemize}

\vspace{3pt}
\noindent \textbf{Human Annotation}: \textbf{3.0}

\vspace{3pt}
\noindent\textbf{Response of Claude-3.7-Sonnet:} \{Rating: \textbf{2}\}. The assistant gives irrelevant and incorrect advice about computer virus prevention and lack accuracy, relevance, and conversational engagement, failing to address the user's real cybersecurity concerns. The low score reflects poor performance, especially in providing safe and useful guidance on a critical topic.

\vspace{3pt}
\noindent\textbf{Response of Qwen2.5-7B-Instruct:} \{Rating: \textbf{5}\}. The assistant meets some standards, but there are still significant deficiencies overall. The assistant's response lacks accuracy and relevance in addressing the human's question about preventing computer viruses. Instead of providing useful advice such as keeping software updated, using antivirus software, and avoiding suspicious websites, the assistant gives vague and unrelated suggestions. The assistant also fails to maintain a conversational and interactive tone, and does not show personalization or creativity in the response.

\vspace{3pt}
\noindent\textbf{Score of \OURS{}:} \textbf{0.326} (Our model directly outputs a dialogue quality score and normalizes it to the [0, 1] range. Compared to other models (Claude-3.7-Sonnet, Qwen2.5-7B-Instruct, etc.), \OURS{}’s predictions are closer to human ratings.)

\end{tcolorbox}

\begin{tcolorbox}[colback=gray!3, colframe=black, arc=2mm, left=3pt,right=3pt,
    boxsep=5pt, boxrule=0.5pt, colframe=black!70, fonttitle=\bfseries, title = A Case of Pairwise Comparison on Daily-MTD-Pair Dataset, breakable]

\vspace{3pt}
\noindent\textbf{Conversation:}
\begin{itemize}
  \item \textbf{Human:} Hi, could you tell me what time it is now?
    \begin{itemize}
      \item \textbf{Assistant A:} The current time is 2:50 AM.
      \item \textbf{Assistant B:} I'm sorry, but I can't provide real-time information.
    \end{itemize}

  \item \textbf{Human:} Ah, it’s so late! I’m about to fall asleep.
    \begin{itemize}
      \item \textbf{Assistant A:} You should rest earlier; maintaining your health is of utmost importance.
      \item \textbf{Assistant B:} Get some rest! A good night's sleep will help you tomorrow.
    \end{itemize}

  \item \textbf{Human:} Yeah, you're right, I know. But I've been really busy with work these days and have been staying late to finish up.
    \begin{itemize}
      \item \textbf{Assistant A:} Work is important; however, one must not neglect their health. Perhaps you could consider arranging your time more effectively and adjusting your schedule accordingly.
      \item \textbf{Assistant B:} It’s important to balance work and rest. Try setting a schedule to avoid overworking—your health comes first!
    \end{itemize}

  \item \textbf{Human:} Yeah, you make a good point. I'll keep that in mind, thanks for the reminder!
    \begin{itemize}
      \item \textbf{Assistant A:} You're welcome. Wishing you health and happiness every day.
      \item \textbf{Assistant B:} You're welcome! Take care of yourself and sleep well.
    \end{itemize}
\end{itemize}

\vspace{3pt}
\noindent \textbf{Human Annotation}: \textbf{A}

\vspace{3pt}
\noindent\textbf{Response of GPT-4o:} ...... So my final decision is: \{"Overall": "\textbf{Fair}"\}. Both Assistant A and Assistant B offer helpful and empathetic responses throughout the conversation. They both encourage the user to prioritize rest and health, with Assistant A focusing on time management and Assistant B emphasizing rest.

\vspace{3pt}
\noindent\textbf{Response of Llama-3.1-8B-Instruct:} ...... So my final decision is: \{"Overall": "\textbf{B}"\}. Assistant B's responses are more concise and to the point, and its tone is more empathetic and encouraging, while Assistant A's responses are slightly longer and more wordy, and its tone is more direct and advisory.
Overall, while both assistants provide helpful and relevant responses, Assistant B's responses are more concise, empathetic, and actionable, making it a slightly better performance.

\vspace{3pt}
\noindent\textbf{Response of \OURS{}:} \textbf{A} (Score A: 0.723, Score B: 0.674, the tie-breaking threshold is 0.01, so A is better, which agrees with human judgment).

\end{tcolorbox}


\begin{tcolorbox}[colback=gray!3, colframe=black, arc=2mm, left=3pt,right=3pt,
    boxsep=5pt, boxrule=0.5pt, colframe=black!70, fonttitle=\bfseries, title = A Case of Multi-Dimensional Comparison on Daily-MTD-Dim Dataset, breakable]

\vspace{3pt}
\noindent\textbf{Conversation:}
\begin{itemize}

  \item \textbf{Human:} Can you give me some travel inspiration?
    \begin{itemize}
      \item \textbf{Assistant A:} You may choose to explore mysterious locations. For example, in South America, there is the breathtaking Machu Picchu in Peru.
      \item \textbf{Assistant B:} How about exploring Japan’s cherry blossoms, Iceland’s northern lights, or Italy’s Amalfi Coast? Adventure awaits!
    \end{itemize}

  \item \textbf{Human:} Can you tell me more about Machu Picchu?
    \begin{itemize}
      \item \textbf{Assistant A:} When one stands atop Machu Picchu, one is awed by this ancient relic. It is a ruin of the Inca Empire, regarded as a dwelling place of the gods. Here, one can gain insights into Inca culture and history while also appreciating the breathtaking scenery.
      \item \textbf{Assistant B:} Machu Picchu is an ancient Incan city perched high in Peru's Andes Mountains. Built in the 15th century, this UNESCO World Heritage site features incredible stone architecture, terraced fields, and breathtaking mountain views. It's best visited during dry season (May-October).
    \end{itemize}

  \item \textbf{Human:} Can you tell me how to get to Machu Picchu?
    \begin{itemize}
      \item \textbf{Assistant A:} You may first fly to Lima, the capital of Peru, and then take either a flight or a bus to Cusco. Finally, you can reach Machu Picchu by train or on foot. This journey presents significant challenges; however, it is also an incredibly rewarding experience.
      \item \textbf{Assistant B:} To reach Machu Picchu, fly to Cusco, Peru, then take a train or hike the Inca Trail to Aguas Calientes. From there, a short bus ride or a steep hike leads to the site. Book tickets and permits in advance for a smooth journey.
    \end{itemize}
\end{itemize}

\vspace{3pt}
\noindent \textbf{Human Annotation}: \{
     Accuracy: Fair, Logicality: Fair, Conversationality: Fair, Relevance: A, 
     Personalization: Fair, Creativity: Fair, Interactivity: A, Emotionality: Fair,
     Informativeness: Fair, Safety: Fair\}
     
\vspace{3pt}
\noindent\textbf{Response of Grok-3:} \{
     Accuracy: Fair, Logicality: Fair, Conversationality: Fair, Relevance: B, 
     Personalization: Fair, Creativity: Fair, Interactivity: A, Emotionality: A,
     Informativeness: B, Safety: Fair\}

\vspace{3pt}
\noindent\textbf{Response of Prometheus-7B:} \{
     Accuracy: A, Logicality: Fair, Conversationality: A, Relevance: A, 
     Personalization: Fair, Creativity: B, Interactivity: A, Emotionality: B,
     Informativeness: A, Safety: Fair\}

\vspace{3pt}
\noindent\textbf{Response of \OURS{}:} \{
     Accuracy: Fair, Logicality: Fair, Conversationality: Fair, Relevance: A, 
     Personalization: Fair, Creativity: Fair, Interactivity: Fair, Emotionality: Fair,
     Informativeness: Fair, Safety: Fair\}

\end{tcolorbox}

\textbf{Remark:} For multi-dimensional comparison tasks, our model achieves alignment with human expert annotations in 9 out of 10 fine-grained dimensions, demonstrating a high degree of consistency. In comparison, Grok-3 aligns with human ratings in 7 dimensions, while Prometheus-7B matches in only 5 dimensions.

\clearpage

\end{document}